\documentclass[11pt,letterpaper]{article}
\usepackage{emnlp2017}
\usepackage{times}
\usepackage{latexsym}
\usepackage{multicol}

\usepackage{pifont}
\usepackage{amsmath}
\usepackage{amsfonts}
\usepackage{amssymb}
\usepackage{hyperref}
\usepackage[font={footnotesize}]{caption}

\newcommand\EE{\mathbb{E}}

\newcommand\PP{\mathbb{P}}

\newcommand{\cmark}{\ding{51}}
\newcommand{\xmark}{\ding{55}}

\def \samooo{\begin{tabular}{@{}c@{}} 
\texttt{???ccccccccccccccccccccccccccccc} \\ 
\texttt{\&\&;?x??????++++++?+?+?++++++++++  } \\ 
\texttt{?vVVV5--5-?-?-?-?-?-?-?s?-ss\{6? } 
\end{tabular}}

\def \samolo{\begin{tabular}{@{}c@{}}
\texttt{??nnnnnnnnnnnnnnnnnneeeee mfe mf} \\ 
\texttt{rerrrrrrrr e an e ao e a  e ho e} \\ 
\texttt{"h"p t t t t t t t ' t h e e a  } 
\end{tabular}}

\def \samoll{\begin{tabular}{@{}c@{}}
\texttt{The increase is a bilday in the } \\ 
\texttt{Sment used a last give you last } \\ 
\texttt{She was the intervice is orced t} 
\end{tabular}}

\def \samloo{\begin{tabular}{@{}c@{}}
\texttt{1x?????????????	????????????????} \\ 
\texttt{Bonererennerere ?Sh???????orann } \\ 
\texttt{unngenngHag  g g?e?????????????	} 
\end{tabular}}

\def \samllo{\begin{tabular}{@{}c@{}}
\texttt{The prope prof ot prote was the } \\ 
\texttt{Wy rronsy ales ale a Claie of th} \\ 
\texttt{Price was one of the plaids rom } 
\end{tabular}}

\def \samlll{\begin{tabular}{@{}c@{}}
\texttt{Republicans friends like come ti} \\ 
\texttt{Researchers have played people a} \\ 
\texttt{The Catalian Office of the docum} 
\end{tabular}}

\def \convconv{\begin{tabular}{@{}c@{}}
\texttt{Official marth Damilicon was eng} \\ 
\texttt{The later , trading touse of the} \\ 
\texttt{First killed sye of Nondon , and} 
\end{tabular}}

\def \samlllsf{\begin{tabular}{@{}c@{}}
\texttt{Marks live up in the club comes the handed up moved to a brief d} \\ 
\texttt{The man allowed that about health captain played that alleged to} \\ 
\texttt{If you have for the past said the police say they goting ight n} \\
\texttt{However , he 's have constance has been apparents are about home} \\
\texttt{The deal share is dipled that a comments in Nox said in one of t} \\
\texttt{Like a sport released not doing the opposition overal price tabl} 
\end{tabular}}

\makeatletter
\def\hlinewd#1{%
  \noalign{\ifnum0=`}\fi\hrule \@height #1 \futurelet
   \reserved@a\@xhline}
\makeatother

\emnlpfinalcopy

\title{Language Generation with Recurrent Generative Adversarial Networks without Pre-training}

\author{
\textbf{Ofir Press}\thanks{~~\scriptsize Denotes equal contribution. Author ordering determined by coin flip.}~$^{1}$,
\textbf{Amir Bar}$^{*1,2}$,
\textbf{Ben Bogin}$^{*1,3}$ \\
\textbf{Jonathan Berant}$^{1}$, \textbf{Lior Wolf}$^{1,4}$ \\
	    $^1$ School of Computer Science, Tel-Aviv University\\
        $^2$  Zebra Medical Vision ~~~ $^3$  IBM Research ~~~       $^4$  Facebook AI Research \\
	    {\tt ofir.press@cs.tau.ac.il} \\
 }

\date{}

\begin{document}

\maketitle

\begin{abstract}
 Generative Adversarial Networks (GANs) have shown great promise recently in image generation. Training GANs for language generation has proven to be more difficult, because of the non-differentiable nature of generating text with recurrent neural networks. Consequently, past work has either resorted to pre-training with maximum-likelihood or used convolutional networks for generation.
In this work, we show that recurrent neural networks can be trained to generate text with GANs from scratch using curriculum learning, 
by slowly teaching the model to generate sequences of increasing and variable length.
We empirically show that our approach vastly improves the quality of generated sequences compared to a convolutional baseline.
\footnote{\scriptsize Code for our models and evaluation methods is available at \url{https://github.com/amirbar/rnn.wgan}}
\end{abstract}

\section{Introduction}

Generative adversarial networks~\cite{GAN} have achieved state-of-the-art results in image generation~\cite{GAN,DCGAN, WGAN,improvedWGAN}. 
For text generation, training GANs with recurrent neural networks (RNNs) has been more challenging, mostly due to the non-differentiable nature of generating discrete symbols. Consequently, past work on using GANs for text generation has been based on pre-training~\cite{seqgan, advDialogue,nmtGAN,nmtGAN2,rttGAN,zhangGenerating,captionGAN} or joint training~\cite{profF,mladGAN} of the generator and discriminator with a supervised maximum-likelihood loss.

Recently, two initial attempts to generate text using purely generative adversarial training were conducted by~\citet{improvedWGAN} and~\citet{boundarySeekingGAN}. 
In these works, a convolutional neural network (CNN) was trained to produce sequences of 32 characters. This CNN architecture is fully differentiable, and the authors demonstrated that it generates text at a reasonable level. However, the generated text was still filled with spelling errors and had little coherence. RNNs are a more natural architecture for language generation, since they condition each generated character on the entire history, and are not constrained to generating a fixed number of characters. 

In this paper, we extend the setup of~\citet{improvedWGAN} and present a method for generating text with GANs. Our main contribution is a model that employs an RNN for both the generator and discriminator, similar to current state-of-the-art approaches for language generation~\cite{sutskever2011generating,mikolov2012statistical,jozefowicz2016exploring}.
We succeed in training the model by using curriculum learning \cite{elman1993learning,bengio2009curriculum,ranzato2015sequence}:
At each stage we increase the maximal length of generated sequences, and train over sequences of variable length that are shorter than that maximal length. In addition,
we aid the model by feeding it with ground truth characters before generation.
We show that these methods vastly improve the quality of generated sequences. Sequences contain substantially more n-grams from a development set compared to those generated by a CNN, and generation generalizes to sequences that are longer than the sequences the model was trained on.

\section{Motivation}

While models trained with a maximum-likelihood objective (ML) have shown success in language generation~\cite{sutskever2011generating,mikolov2012statistical,jozefowicz2016exploring}, there are drawbacks to using ML, that suggest training with GANs. First, using ML suffers from ``exposure bias'', that is, at training time the model is exposed to gold data only, but at test time it observes its own predictions, and thus wrong predictions quickly accumulate, resulting in bad text generation. 

Secondly, the ML loss function is very stringent. When training with ML, the model aims to allocate all probability mass to the $i$-th character of the training set given the previous $i-1$ characters, and considers any deviation from the gold sequence as incorrect, although there are many possible sequences given a certain prefix. GANs suffer less from this problem, because the objective is to fool the discriminator, and thus the objective evolves dynamically as training unfolds. While at the beginning the generator might only generate sequences of random letters with spaces, as the discriminator learns to better discriminate, the generator will evolve to generate words and after that it may advance to longer, more coherent sequences of text. This interplay between the discriminator and generator helps incremental learning of text generation. 

\section{Preliminaries} \label{sec:preliminaries}
\citet{improvedWGAN} and~\citet{boundarySeekingGAN} trained a purely generative adversarial model (without pre-training) for character-level sentence generation. 
We briefly review the setup of~\citet{improvedWGAN},  who use the Improved Wasserstein GAN objective~\cite{WGAN,improvedWGAN}, which we employ as well.~\citet{boundarySeekingGAN} have a similar setup,  but employ the Boundary-Seeking GAN objective.

The generator $G$ in~\citet{improvedWGAN} is a CNN that transforms a noise vector $z \sim N(0,1)$ into a matrix $M \in \mathbb{R}^{32 \times V}$, where $V$ is the size of the character vocabulary, and 32 is the length of the generated text. In this matrix the $i$-th row is a probability distribution over characters that represents a prediction for the $i$-th output in the character sequence. To decode a sequence, they choose the highest probability character in each row.
The discriminator $D$ is another CNN that receives a matrix as input and needs to determine if this matrix is the output of the generator $G$ or sampled from the real data (where each row in the matrix now is a one-hot vector).
The loss of the Improved WGAN generator is:
$$L_G = -\EE_{\widetilde{x} \sim \PP_g}[D(\widetilde{x})],$$
and the loss of the discriminator is:
\begin{multline*} 
L_D = 
\EE_{\widetilde{x} \sim \PP_g}[D(\widetilde{x})] - \EE_{x \sim \PP_r} [D(x)]
 \\
+ \lambda \EE_{\widehat{x} \sim \PP_{\widehat{x}}} [(\| \nabla_{\widehat{x}} D(\widehat{x})   \|_2 - 1)^2],
\end{multline*}
Where $\PP_r$ is the data distribution and $\PP_g$ is the generator distribution implicitly defined by $\widetilde{x} = G(z)$.
The last term of the objective controls the complexity of the discriminator function and penalizes functions that have high gradient norm, that is, change too rapidly.
$\PP_{\widehat{x}}$ is defined by sampling uniformly along a straight line between a point sampled from the data distribution and a point sampled from the generator distribution. 

A disadvantage of the generators in~\citet{improvedWGAN} and~\citet{boundarySeekingGAN} is that they use CNNs for generation, and thus the $i$-th generated character is not directly conditioned on the entire history of $i-1$ generated characters. This might be a factor in the frequent spelling mistakes and lack of coherence in the output of these models. We now present a model for language generation with GANs that utilizes RNNs, which are state-of-the-art in language generation. 

\begin{table*}[]
\centering
\footnotesize
\caption{ Samples and evaluation of the baseline model from~\protect\citet{improvedWGAN}. }
\label{results-baseline}
%\vskip 0.1in
\begin{tabular}{lllll}
\hline
 Samples & \multicolumn{4}{c}{\%-IN-TEST-$n$} \\ 
  & 1 & 2& 3& 4    \\ \hline
\convconv & 64.4 & 25.9 & 5.1 & 0.4 \\ \hline
\end{tabular}
%\vskip -0.1in
\end{table*}

\section{Recurrent Models}
We employ a GRU~\cite{gru} based RNN for our generator and discriminator. 
The generator is initialized by feeding it with a noise vector $z$ as the hidden state, and an embedded start-of-sequence symbol as input.
The generator then generates a sequence of distributions over characters, using a softmax layer over the hidden state at each time step.

Because we want to have a fully-differentiable generator, 
the input to the RNN generator at each time step is not the most probable character from the previous time step. Instead we employ a continuous relaxation, and provide at time step $i$ the weighted average representation given by the output distribution of step $i-1$. More formally, let $\alpha_{i-1}^c$ be the probability of generating the character $c$ computed at time step $i-1$, and let $\phi(c)$ be the embedding of the character $c$, then the input to the GRU at time step $i$ is $\sum_c \alpha_{i-1}^c \phi(c)$. This is fully differentiable compared to $\arg\max_{\phi(c)} \alpha_{i-1}^c$. We empirically observe that the RNN quickly learns to output very skewed distributions.

The \textbf{discriminator} is another GRU that receives a sequence of character distributions as input, either one-hot vectors (for real data) or softer distributions (for generated data). 
Character embeddings are computed from the distributions and fed into the GRU. 
The discriminator then takes the final hidden state and feeds it into a fully connected layer which outputs a single number, representing the score that the discriminator assigns to the input. The models are trained with the aforementioned Improved WGAN objective (Section~\ref{sec:preliminaries}).

An advantage of a recurrent generator compared to the convolutional generator of~\citet{improvedWGAN} and \citet{boundarySeekingGAN} is that can output sequences of varying lengths, as we empirically show in Section~\ref{sec:results}. 

Our baseline model trains the generator and discriminator over sequences of length $32$, similar to how CNNs were trained in~\citet{improvedWGAN}. We found that training this baseline was difficult and resulted in nonsensical text. 
We now present three extensions that stabilize the training process.

\begin{table*}[]\label{tab:table2}
\centering
\footnotesize
\caption{ Samples and evaluation of our models with an RNN generator and discriminator and various extensions. For the CL+VL+TH model we present results for generated sequences of length $32$ and  length $64$.}
\label{results}
%\vskip 0.1in
\begin{tabular}{ccclcccc}
\hline
CL  & VL  & TH & Samples &    \multicolumn{4}{c}{\%-IN-TEST-$n$} \\  
 \multicolumn{4}{c}{} & 1 & 2& 3& 4    \\ \hline
\xmark &  \xmark  & \xmark  &  \samooo & 28.8 & 3.7 & 0.0 & 0.0   \\ \hline
\xmark &  \cmark  & \xmark  &  \samolo & 80.6 & 8.6 & 0.0 & 0.0   \\ \hline
\cmark &  \xmark  & \xmark  &  \samloo & 27.0 & 7.9 & 2.0 & 0.0 \\ \hline
\cmark &  \cmark  & \xmark  &  \samllo & 68.1 & 24.5 & 4.4 & 0.5 \\ \hline
\xmark &  \cmark  & \cmark  &  \samoll & 79.4 & 44.6 & 11.5 & 0.7 \\ \hline
\cmark &  \cmark  & \cmark  &  \samlll & \textbf{87.7} & \textbf{54.1} & \textbf{19.2} & \textbf{3.8} \\ \hlinewd{2pt}
\cmark &  \cmark  & \cmark  &  \textbf{Sequences of length 64. Examples in Table 3.} & 87.5 & 51.3 & 15.1& 1.7 \\ \hline
\end{tabular}

%\vskip -0.1in
\end{table*}

\begin{table*}[]\label{tab:table3}
\centering
\footnotesize
\caption{Samples of length $64$ generated by the CL+VL+TH model.}
\label{results-64}
%\vskip 0.1in
\begin{tabular}{l}
\hline 
 \samlllsf  \\ \hline
\end{tabular}
%\vskip -0.1in
\end{table*}

\paragraph{Curriculum Learning (CL)} 
In this extension, we start by training on short sequences and then slowly increase sequence length. In the first training stage, the generator $G$ generates sequences of length $1$, and the discriminator $D$ receives real and generated sequences of length $1$ as input. Then, the generator  generates sequences of length $2$ and the discriminator receives sequences of length $2$. We increase sequence length in this manner until the maximum length of 32 characters.

\paragraph{Variable Length (VL)}
Here, we define a maximum length $l$, and generate during training sequences of every length $\leq l$ in every batch. Without curriculum learning, this amounts to training $G$ and $D$ in every batch with sequences of length $i$, $1 \leq i \leq 32$.
With curriculum learning, we generate at each step sequences of length $i$, $1 \leq i \leq l$, and slowly increase $l$ throughout training. 

\paragraph{Teacher Helping (TH)} 
Finally, we propose a procedure where we help the generator learn to generate long sequences by conditioning on shorter ground truth sequences. Recall that in our baseline, the generator generates an entire sequence of characters that are fed as input to the discriminator. Here, when generating sequences of length $i$, we feed the generator a sequence of $i-1$ characters, sampled from the real data. Then, the generator generates a distribution over characters for the final character, which we concatenate to the real characters and feed as input to the discriminator. The discriminator observes a sequence of length $i$ composed of $i-1$ real characters and one character that is either real or generated.
This could be viewed as a conditional GAN~\cite{CGAN}, where the first $i-1$ characters are the input and the final character is the output. Note that this extension may suffer from exposure bias, similar to the ML objective, and we plan to address this problem in future work.

\section{Results} \label{sec:results}
To directly compare to~\citet{improvedWGAN}, we follow their setup and train our models on the Billion Word dataset~\cite{1billion}. 
We evaluate by generating 640 sequences from each model and measuring \%-IN-TEST-$n$, that is, the proportion of word $n$-grams from generated sequences that also appear in a held-out test set. We evaluate these metrics for $n \in \{1, 2, 3,4\}$. 
Our goal is to measure the extent to which the generator is able to generate real words with local coherence. %, that have not been observed at training time.

In contrast to~\citet{WGAN} and~\citet{improvedWGAN}, where the generator is trained once for every $10$ training iterations of the discriminator, we found that training the generator for $50$ iterations every $10$ training iterations of the discriminator resulted in superior performance. In addition, instead of using noise vectors sampled from the $N(0,1)$ distribution as in~\citet{improvedWGAN}, we sample noise vectors from the $N(0,10)$ distribution, since we found this leads to a greater variance in the generated samples when using RNNs. 

In all our experiments, we used single layer GRUs for both the discriminator and generator. The embedding dimension  and hidden state dimension are both of size $512$.  

Following~\citet{improvedWGAN}, we train all our models on sequences whose maximum length is $32$ characters.
Table 1 shows results of the baseline model of~\citet{improvedWGAN}, and Table 2 presents results of our models with various combinations of extensions (Curriculum Learning, Variable Length, and Teacher Helping).  Our best model combines all of the extensions and outperforms the baseline by a wide margin on all metrics. 

The samples show that models that used both the Variable Length and Teacher Helping extensions performed better than those that did not. This is also backed by the empirical evaluation, which shows that 3.8\% of the word 4-grams generated by the CL+VL+TH model also appear in the held-out test set.  
The weak performance of the curriculum learning model without the other extensions shows that curriculum learning by itself does not lead to better performance, and that training on variable lengths and with Teacher Helping is important. 
We note that curriculum learning did not perform well at generating sequences of length $32$, but did perform well at generating sequences of shorter lengths earlier in the training process. For example, the model that used only curriculum learning had a \%-IN-TEST-1 of $79.9$ when it was trained on sequences of length $5$. This decreased to $59.7$ when the model reached sequences of length $10$, and continued decreasing until training stopped. This also shows the importance of Variable Length and Teacher Helping. 

Finally, to check the ability of our models to generalize to longer sequences, we generated sequences of length $64$ with our CL+VL+TH model, which was trained on sequences of up to $32$ characters (Table 3).  We then evaluated the generated text, and this evaluation shows that there is a small degradation in performance (Table 2).

\section{Conclusion}
We show for the first time an RNN trained with a GAN objective that learns to generate natural language from scratch. Moreover, we demonstrate that our model generalizes to sequences longer than the ones seen during training. In future work, we plan to apply these models to tasks such as image captioning and translation, comparing them to models trained with maximum likelihood.

\bibliography{emnlp2017}
\bibliographystyle{emnlp_natbib}

\end{document}